\documentclass{article}

\usepackage{xarxiv}

\usepackage[utf8]{inputenc} 
\usepackage[T1]{fontenc}    
\usepackage{hyperref}       
\usepackage{url}            
\usepackage{booktabs}       
\usepackage{amsfonts}       
\usepackage{nicefrac}       
\usepackage{microtype}      
\usepackage{lipsum}
\usepackage{epsfig,amsmath,amssymb,color,url}

\usepackage[numbers,sort&compress]{natbib}

\renewcommand{\cite}[1]{\citep{#1}}
\renewcommand{\citet}[1]{\citeauthor{#1} \citep{#1}}
\definecolor{mycolor}{rgb}{0,0.4,0.6}
\definecolor{mycolor}{gray}{0}

\usepackage{framed}
\usepackage{xcolor}
\colorlet{shadecolor}{gray!10} 
\usepackage{mdframed}
\usepackage{colortbl}
\usepackage{multicol}

\definecolor{dblu}{RGB}{0, 79,124}             
\definecolor{grey}{RGB}{146, 146, 146}     
\definecolor{blu}{RGB}{0,141,196}
\definecolor{lightblue}{RGB}{0,200,250}
\definecolor{hicolor}{RGB}{30,30,30} 

\newcommand{\frablu}[1]{\begin{mdframed}[linewidth=1,linecolor=mycolor,backgroundcolor=blu!5]{#1} \end{mdframed}}
\newcommand{\fre}{\end{mdframed}}

\renewcommand{\vec}[1]{\boldsymbol{#1}}

\newcommand{\haty}{\hat{y}}

\definecolor{cellcolorb}{rgb}{0,0.4,0.6}
\definecolor{cellcolorc}{rgb}{0,0.4,0.6}
\definecolor{cellcolora}{rgb}{1,1,1}

\newcommand{\tx}[1]{& }

\newcommand{\Prob}{P}
\newcommand{\prob}{p}

\newcommand{\argmin}{\operatorname*{argmin}}

\newcommand{\given}{\, | \,}

\renewcommand{\vec}[1]{#1}
\newcommand{\cX}{\mathcal{X}}
\newcommand{\cY}{\mathcal{Y}}
\newcommand{\cH}{\mathcal{H}}

\begin{document}

\title{Prescriptive Machine Learning for Automated Decision Making: Challenges and Opportunities}

\author{
  Eyke H\"ullermeier \\
  Institute of Informatics\\
  University of Munich (LMU)\\
  \texttt{eyke@lmu.de} 
}

\maketitle

\begin{abstract}
Recent applications of machine learning (ML) reveal a noticeable shift from its use for \emph{predictive} modeling in the sense of a data-driven construction of models mainly used for the purpose of prediction (of ground-truth facts) to its use for \emph{prescriptive} modeling. What is meant by this is the task of learning a model that stipulates appropriate decisions about the right course of action in real-world scenarios: Which medical therapy should be applied? Should this person be hired for the job? As argued in this article, prescriptive modeling comes with new technical conditions for learning and new demands regarding reliability, responsibility, and the ethics of decision making. Therefore, to support the data-driven design of decision-making agents that act in a rational but at the same time responsible manner, a rigorous methodological foundation of prescriptive ML is needed. The purpose of this short paper is to elaborate on specific characteristics of prescriptive ML and to highlight some key challenges it implies. Besides, drawing connections to other branches of contemporary AI research, the grounding of prescriptive ML in a (generalized) decision-theoretic framework is advocated.


\end{abstract}

\keywords{Predictive modeling \and Prescriptive modeling \and Decision theory \and Bounded rationality \and Uncertainty \and Preference learning \and Counterfactual reasoning \and AI ethics \and Fairness }

\section{Introduction}

Machine learning (ML) methodology, fueled with access to ever-increasing masses of data and unprecedented computing power, has been the main driving factor of recent progress in artificial intelligence (AI) and its applications in various branches of science and technology, industry and business, economics and finance, amongst others. In this regard, ML is most commonly perceived as a means for \emph{predictive modeling}, that is, for the data-driven construction of a model that is mainly used for the purpose of predicting unknown facts in a specific context\,---\,albeit models may, of course, serve other purposes, too, such as understanding and explanation, or may have a more descriptive flavor. A predictive model, or ``predictor'' in ML jargon, is trained in a supervised manner on cases encountered by the ``learner'' over the course of time, such as emails categorized as spam or non-spam, and the model is then used to make predictions in future situations, \textit{e.g.}, to automatically mark new emails.

Looking at emerging applications of ML methodology, there is a visible shift from predictive modeling to \emph{prescriptive modeling}, by which we mean the task of learning a model that stipulates appropriate decisions or actions to be taken in real-world scenarios. In fact, decisions are nowadays increasingly automated and made by algorithms instead of humans, and most of these \emph{automated decision making} (ADM) algorithms are trained on data using ML methods. For example, think of decisions in the context of employees recruitment, such as hiring or placement decisions  \cite{pess_er20}. Another important case is the data-driven construction of individualized treatment rules in personalized medicine \cite{zhao_ei12}, and further examples can be found in other domains, including jurisdiction \citep{klein_hd18}, finance \citep{hans_tv20}, and disaster management \citep{zago_dm13}. The distinction between predictive and prescriptive modeling can be compared with the distinction between predictive (``What will happen?'') and prescriptive (``How to make something happen?'') tasks as commonly made in business analytics \cite{bert_tp20}. 

As shown by the above examples of ADM, machine learning is gaining in societal significance and increasingly impacting on our daily life, a development that comes with increasing demands for ML methodology:
``prescriptive ML'' is supposed to (\textit{i}) stipulate decisions in a \emph{transparent}, \emph{principled} (rather than ad hoc), and \emph{ethically justifiable} manner, and (\textit{ii}) to be sufficiently aware of its own (in)competence, notably its (un)certainty about the right course of action, so as to assure its \emph{reliability} and \emph{responsibility} \citep{lepr_ft18,cast_ua19}.

In addition to raising new demands, prescriptive modeling also exhibits notable technical differences compared to standard ML settings. In particular, the key assumption of an objective ``ground truth'' underlying every prediction, revealed to the learner as training information and serving as a reference to evaluate the correctness of predictions, is no longer justified. For example, there is nothing like a ground-truth therapy plan for a patient, a ground-truth dietary plan for an athlete preparing for a competition, or a ``true'' abstract of an article in automatic text summarization \cite{gamb_ra17}. Even when taking the stance that \emph{optimality} may serve as a surrogate for truth, an optimal therapy or plan could never be determined or verified, as it would require counterfactual reasoning about (perhaps infinitely many) ``possible worlds'' (\textit{e.g.}, how the patient would have developed under each possible therapy, or the athlete would have performed under each possible dietry) \cite{pros_ci20}. 

As detailed further below, prescriptive ML also exhibits other characteristics distinguishing it from predictive ML, and showing that standard ML methodology does not fully comply with the demands and technical subtleties of prescriptive modeling. 
What is needed, therefore, is an ML methodology specifically tailored to the task of prescriptive modeling.


\section{Predictive versus Prescriptive ML}

Predictive modeling aims to capture the (stochastic) dependence between \emph{instances} $X$ and associated \emph{outcomes} $Y$. In the standard setting of supervised learning, training data consists of a set $\mathcal{D}$ of examples in the form of tuples $(\vec{x}, y) \in \mathcal{X} \times \mathcal{Y}$, 
with $\mathcal{X}$ being the instance space and $\mathcal{Y}$ the set of possible outcomes. Typically, these examples are assumed to be independent and identically distributed (i.i.d.) according to some unknown probability measure $\Prob$ on $\mathcal{X} \times \mathcal{Y}$ (with mass or density function $\prob$), which is the target of \emph{generative} learning. \emph{Discriminative} learning proceeds from a \emph{hypothesis space} $\mathcal{H} \subset \cY^\cX$, where each hypothesis is a map (predictor) $\cX \longrightarrow \cY$, and a loss function $\ell: \, \mathcal{Y} \times \mathcal{Y} \longrightarrow \mathbb{R}$. The learner then seeks the ``best predictor'', \emph{viz.}\ the expected loss minimizer (Bayes predictor)
\begin{equation}\label{eq:bayespred}
h^* \in \argmin_{h \in \cH} \int_{\cX \times \cY} \ell( h(\vec{x}) , y) \, d \, \Prob  \, ,
\end{equation}
and leverages the data $\mathcal{D}$ to produce an approximation $h$ thereof. The latter can then be used to make predictions $\haty_{q} = h(x_q)$ for new query instances $\vec{x}_{q} \in \cX$. 

A basic assumption in this setting is that $Y$ does (or will) exist independently of the prediction. In other words, for every concrete instance $X$, there is a ``ground truth'' $Y$, and hence a ground-truth (albeit non-deterministic) dependence between instances and outcomes (specified by the map $X \mapsto \prob(Y \given X)$). For example, if the prediction is a weather forecast for tomorrow, based on weather conditions in the last days ($X$), the ``ground-truth weather'' $Y$ does (or will) indeed exist, independently of the prediction, just like a disease exists independently of the diagnosis or a handwritten digit independently of its image-based prediction. 


This view of outcomes $Y$ or, more generally, a dependence between instances and outcomes, as ``ontic'' entities that ought to be ``discovered'', and hence of ML as an \emph{analytic} task, is arguably less appropriate in the context of ADM, where $Y$ is a decision made by the learner in a situation $X$. Here, the problem of constructing a decision model is essentially of \emph{synthetic} nature. Imagine, for example, that $Y$ is not a diagnosis but a drug or a therapy. In this case, $Y$ is not an ontic entity and should be perceived as a \emph{prescription} rather than a prediction. As a consequence, a real ground truth may not exist in prescriptive ML. 
Moreover, as explained above, the notion of ``best decision'' is not a good surrogate of a ground truth either, as it is counterfactual and cannot be determined in practice, especially not retrospectively. In fact, by taking actions in the real world, a prescriptive learner is \textit{influencing and possibly changing} the world through intervention, 
in contrast to predictive modeling, where the learner is \textit{reasoning about} the world from an outside perspective (albeit an indirect influence is possible if decisions are made by others based on the learner's predictions \cite{bert_tp20,elma_dt20}). Consequently, the goal of learning shifts from targeting a ground truth to finding a ``practicable'' decision model, where practicability will typically refer to several criteria simultaneously.

Thus, while predictive modeling is essentially an analytic task, prescriptive modeling is of synthetic nature. Instead of hypothesizing about an underlying ground truth, which may not exist or at least be unavailable for training and evaluation, prescriptive ML is seeking practicable decision models that balance value, complexity, cost, fairness, etc. in an appropriate manner.
These differences between predictive and prescriptive ML have implications for the formalization of ML problems and pose several challenges for the methodology needed to tackle them. 

\begin{itemize}
\item
\textbf{Challenge 1: Weak supervision.}
Training information in prescriptive ML will typically be weak, biased, incomplete, and possibly context-dependent, thereby posing severe challenges for learning. This is especially true for observational data collected from human DMs acting as ``teachers'' \cite{atan_lo18,fern_ov19}. In fact, decisions made in the past are presumably suboptimal and likely to be biased toward the teacher's preferences, so that, in contrast to supervised learning, observed decisions cannot be considered as ground-truth examples suggesting the best course of action in a given context. Besides, training information will often be incomplete, lacking information about ``counterfactuals'', \textit{i.e.}, outcomes for those decisions that were not made, and hence raising the need for counterfactual reasoning \cite{kall_rp17}. At best, a decision is accompanied by some hint at how effective it was, or perhaps by a relative comparison with another decision\,---\,for example, if one candidate $A$ failed to accomplish a job while a second one $B$ succeeded, this could be captured in terms of a preference $B \succ A$.
Finally, decisions may exhibit various types of dependencies. Often, for example, decisions are not made individually but collectively, \textit{i.e.}, in the context of a set of other problems. For instance, the decision of whether or not a candidate is accepted for a study program with limited capacity depends on which other candidates also applied. Obviously, standard assumptions on the data-generating process as commonly made in the predictive modeling setup, such as i.i.d.\ observations, are violated in such cases. 

\item
\textbf{Challenge 2: Complex prescriptions and performance criteria.}
Decisions are not always atomic units but may possess an internal structure, so that decision spaces are structured spaces. For example, the structure can be combinatorial if a (global) decision comprises several sub-decisions \cite{chev_ph08,drag_cp18}, like in a multi-step therapy, where the doctor may need to decide about the type of medication, the dosages, and the time in-between the administrations.
Moreover, in cases where the learner is only supposed to make recommendations, while leaving the final course of action to the user, the prescription of a sole decision may not provide optimal support. Instead, a recommendation might be a ``pre-prescription'' in the form of a \emph{set} of promising decisions, \textit{i.e.}, a subset of candidates the user can choose from \cite{mpub420}\,---\,including complete abstention in the extreme case. Or, the candidates could be further  prioritized by returning a ranking instead of an unsorted set. Regarding prescriptive performance, the outcome of a decision might be rated in terms of \emph{multiple performance criteria} (\textit{e.g.}, medicinal efficacy of a therapy, side-effects, cost, etc.) at the same time, which may lead to incomparability (or weaker comparability, for example, Pareto-dominance) and further increase the complexity of the prescription task \cite{blas_mc07,corr_ro13}.


\item
\textbf{Challenge 3: Uncertainty representation.}
The notion of uncertainty is of utmost importance and plays a fundamental role in prescriptive modeling\,---\,see \textit{e.g.}\  \cite{grot_ot20} for a discussion from an ethical and epistemological perspective. First of all, the learner's  uncertainty-awareness and the adequate representation of its ``prescriptive uncertainty'' is a key prerequisite for responsible and reliable decision making, including the possibility to delay or abstain from a decision if uncertainty is too high. Deciding about the right course of action will require a distinction between different types and sources of uncertainty, notably between \emph{aleatoric} and \emph{epistemic} uncertainty, whence uncertainty representation should go beyond standard probabilistic models \cite{mpub440}. Likewise, a proper representation of uncertainty is essential for the two main components of a  decision-theoretic framework as advocated in Section \ref{sec:dtf}, namely preference and uncertainty (about outcomes of decisions in a given context), because these are learned from data and hence hypothetical; again, a distinction between aleatoric and epistemic uncertainty is important in this regard, with the latter essentially representing uncertainty about uncertainty.   

\item
\textbf{Challenge 4: Complexity and resource constraints.} 
Facing a model construction task of synthetic nature, one will naturally strive for a good compromise between value and complexity of a decision model in prescriptive ML. Indeed, 
if the learner has to make decisions under real-world conditions, 
constraints regarding \emph{time and computational resources} as well as the availability of information must be taken into account. 
In fact, depending on the situation and application context, time and computational resources for collecting information and applying decision models might be limited. Moreover, in the context of prescriptive ML, the decision maker is not necessarily a powerful computer but perhaps a human supported by a low-level computing device (\textit{e.g.}, a smartphone). The DM's resources to collect, validate, and enter data might be scarce, or decisions must be made quickly. In the extreme case, instead of running an algorithm on a computer, the decision is made by the human himself\,---\,imagine, \textit{e.g.}, a medical doctor who needs to make decisions in emergency cases. In such situations, mental capacity will only allow for applying a ``rule of thumb''. In any case, the complexity of a decision model should be well adapted to the experience of the DM, the information at hand, the computational resources available, as well as possible time constraints to be met.

\item
\textbf{Challenge 5: Ethics and fairness.} Acting in the world comes with a certain responsibility, especially if decisions have a direct influence on people. Consequently, in addition to rationality, \emph{ethical criteria} such as fairness have a major role to play in prescriptive ML. Broadly speaking, decisions should not only be rational but also fair and ethical. Although the notion of fairness in ML has been studied quite intensively in the recent past, there is still very little work on combining fairness and rationality, for example, by formalizing fairness in a decision-theoretic context \cite{kami_dt12,frie_ti21}.
Moreover, ML algorithms must be generalized so as to allow for learning fair decision models. This will also influence the learning objectives and performance measures. For example, the optimization of \emph{average} or \emph{expected performance} might be inappropriate in a social context, as it may suggest sacrificing performance on an ethnic minority for the benefit of a bigger subpopulation \cite{slow_ab21}.


\end{itemize}


\section{Positioning and Related Topics}

The field of machine learning has grown quickly over the last decades, and a wide repertoire of ML methodologies is now available. Prescriptive ML has connections to many of these methodologies, including weakly supervised learning \citep{zhou_ab18}, preference learning \cite{mpub218}, 
structured output prediction \cite{baki_ps,maes_sp09,dopp_sp14}, as well as uncertainty representation and quantification \cite{mpub272,kend_wu17,abda_ar21,mpub440}.
Besides, as already mentioned, recent work on ethical, legal, and societal aspects of AI
 have recently come to the fore. Especially relevant for prescriptive ML is recent work on \emph{fairness} \citep{shre_fi19,mehr_as21,tava_fc20}, which has mainly been formalized in terms of constraints that the predictions of an ML model should obey \citep{zlio_md17,shre_fi19,pito_fi21,zehl_fi21}.

In terms of methodology, a basic distinction (according to the type of supervision) is often made between the paradigms of supervised, unsupervised, and reinforcement learning \citep{sutt_rl}. Adhering to this rough categorization, prescriptive ML is perhaps best positioned in-between the latter two. 
The type of training data in prescriptive ML resembles the type of feedback in reinforcement learning (RL) and related settings such as (contextual) multi-armed bandits \cite{latt_ba}. With the latter, it also shares the view of a learner as an \emph{acting} rather than only predicting agent. However, prescriptive ML, at least in its basic form, is more geared toward \emph{one-time} rather than sequential decisions, \emph{passive} rather than active, and \emph{offline} (``batch'') rather than online learning, which are properties it has in common with the setting of supervised learning.
Moreover, algorithms for RL or bandits essentially compensate for the lack of counterfactuals through trial and error strategies, often requiring thousands of trials before a good policy is found \cite{mann_ts04}\,---\,the excessive exploration required for optimization (albeit to a lesser extent for \emph{satisficing} \cite{reve_si17}) is not only inappropriate from an ethical point of view but also in conflict with the quest for per-instance guarantees \cite{grot_ot20}.


More relevant from this perspective is the problem of offline (policy) learning from data of the form $(x,a,u)$, where information about the outcome or utility $u$ caused by a decision $a$ made in a specific context $x$ is given, but lacking information about counterfactuals (\textit{i.e.}, outcomes for decisions $a' \neq a$). 
Variants of this problem have recently been considered in different fields. In machine learning, it is also referred to as \emph{learning from logged bandit feedback} or \emph{batch learning from bandit feedback} \cite{beyg_to09}. This setting requires causal inference and counterfactual estimation \cite{pear_cm,bott_cr13}, on the basis of which (supervised) learning principles such as empirical risk minimization can be generalized. For example, \emph{counterfactual risk minimization} has been proposed as a principle for learning in this setting \cite{swam_cr15,swam_ts15}. \emph{Regress-and-compare} approaches \cite{qian_pg11,bert_tp20} train regression models for estimating (counterfactual) outcomes, one per decision, and then do ``estimated utility maximization'' \cite{mpub065}. 
As another approach for dealing with partial (observational) data, the use of \emph{propensity scores} (conditional probability of a treatment/decision given the context features) has received increasing interest in the field of (medical) statistics \cite{rose_tc83}. Combined with weighted classification, it is used for learning individualized treatment rules in personalized medicine \cite{west_ps10,zhou_rw17}. 
Also, adaptations of classification trees for minimizing a prescription error have recently been proposed \cite{kall_rp17,bert_op19}. Related to this is work on causal trees and forests \cite{athe_rp16,wage_ea18,powe_sm17}. The use of so-called \emph{instrument variables} for the purpose of counterfactual prediction, in combination with deep learning methods, is proposed in \cite{hart_di17}. 
Although the problem can be seen as a special case of prescriptive modeling, the setups considered are quite specific and often coming with strong assumptions (\textit{e.g.}, binary treatment case or known propensities/logging policy). Moreover, major challenges regarding reliability, responsibility, and ethics are not addressed.

\section{A Decision-theoretic Approach}
\label{sec:dtf}

The idea of machine learning as a means for automated data-driven decision making suggests a methodology of prescriptive ML grounded in (prescriptive) \emph{decision theory} as a formal foundation of rational decision making under uncertainty. A decision-theoretic grounding promises to support the systematic development of prescriptive ML methods in a principled (rather than ad hoc) manner. Moreover, it involves the elucidation of underlying assumptions, values, and norms, as well as the explication and (axi\-omatic) justification of a decision-theoretic principle, thereby contributing to the transparency and social acceptability of algorithmic decisions \cite{good_eu17}.

\begin{figure*}
\frablu{
\begin{center}
\includegraphics[scale=0.4]{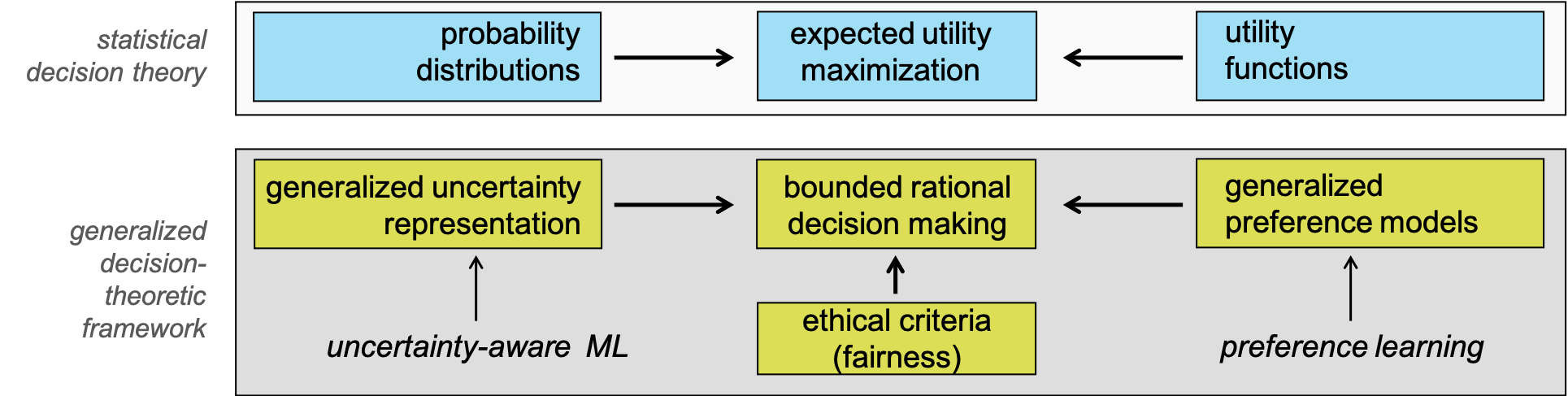} \qquad
\caption{\footnotesize In standard statistical decision theory (above), the decision principle of expected utility maximization takes probability distributions for uncertainty and utility functions for preference representation as input. In the envisioned prescriptive ML methodology (below), prescriptions are produced in the form of bounded rational decisions (middle). The latter are made on the basis of suitably represented preferences (right) and uncertainty (left) in the current context, which are both obtained through machine learning. The methodology shall also accommodate additional ethical criteria (such as fairness), taken as another ``input'' for decision making. 
}

\label{fig:overview2}
\end{center}}
\end{figure*}

Statistical decision theory \cite{ferg_ms,berg_sd80,berg_sd} essentially builds upon the notions of \emph{preference} and \emph{uncertainty}, which are formalized in terms of utility functions and probability distributions, respectively (cf.\ Fig.\ \ref{fig:overview2}, upper panel). While this theory does provide a suitable basis for standard (predictive) ML\,---\,for example, the principle of \emph{expected loss minimization} in ML is in direct correspondence with the celebrated (and axiomatically justified) principle of expected utility maximization \cite{vonn_to,sava_tf}\,---\,utility functions and probability distributions are not expressive enough in the context of prescriptive ML. For example, preferences are not pre-specified but learned from data, and hence uncertain and maybe not readily expressible in numeric form. 
Moreover, the quest for a self-aware decision maker calls for an expressive representation of uncertainty, notably a distinction between epistemic (``I don't know'') and aleatoric (``I know it's a matter of chance'') uncertainty.  
What is arguably needed, therefore, is a novel decision-theoretic foundation of prescriptive modeling built upon recent work on generalized uncertainty representation in ML \cite{mpub440} as well as preference modeling and learning \cite{mpub218}. 

Further, classical decision theory does not account for complexity and computational aspects, which are essential from an ML perspective and may indeed prevent a DM from realizing ``perfect rationality''. Therefore, in addition to uncertainty and preference learning, prescriptive ML should leverage the concept of \emph{bounded rationality} \cite{russ_ra16,whee_br20}, thereby connecting three subfields of modern AI (cf.\ Fig.\ \ref{fig:overview2}, lower panel).  Last but not least, a methodology of prescriptive ML shall also be capable of accommodating  \emph{social and ethical} aspects, especially fairness \citep{shre_fi19,mehr_as21,tava_fc20}, as key concepts of contemporary AI and responsible decision making.

\section{Summary}

We emphasized the need for a sound methodological foundation of prescriptive modeling, that is, the use of machine learning to support automated decision making, and argued that standard ML methodology geared toward predictive modeling does not fully comply with the specific characteristics of prescriptive ML and the challenges it implies. These challenges include the need for learning from weak and possibly biased training information that lacks ``optimal'' decisions provided as reference, the quest for reliable prescriptions with individual guarantees, for responsibility and fairness, as well as an increased need for an adequate representation of uncertainty and for taking resource constraints into consideration.






Viewing prescriptions as (bounded) rational decisions under uncertainty, we advocate a methodology grounded in decision theory and leveraging novel approaches to handling uncertainty and preference in machine learning. A methodology of that kind would support the systematic development of prescriptive ML methods in a principled manner. Moreover, as a decision-theoretic approach involves the elucidation of underlying assumptions, values, and norms, it may contribute to the transparency and social acceptability of algorithmic decisions. As such, it would provide a theoretical foundation for the data-driven design of decision-making agents that act in a rational but at the same time responsible manner, and thereby a profound basis of algorithmic decision making.



\begin{thebibliography}{65}
\providecommand{\natexlab}[1]{#1}
\providecommand{\url}[1]{\texttt{#1}}
\expandafter\ifx\csname urlstyle\endcsname\relax
  \providecommand{\doi}[1]{doi: #1}\else
  \providecommand{\doi}{doi: \begingroup \urlstyle{rm}\Url}\fi

\bibitem[Abdar et~al.(2021)Abdar, Pourpanah, Hussain, Rezazadegan, Liu,
  Ghavamzadeh, Fieguth, Cao, Khosravi, Acharya, Makarenkov, and
  Nahavandi]{abda_ar21}
M.~Abdar, F.~Pourpanah, S.~Hussain, D.~Rezazadegan, L.~Liu, M.~Ghavamzadeh,
  P.~Fieguth, X.~Cao, A.~Khosravi, U.R. Acharya, V.~Makarenkov, and
  S.~Nahavandi.
\newblock A review of uncertainty quantification in deep learning: Techniques,
  applications and challenges.
\newblock \emph{Information Fusion}, 76:\penalty0 243--297, 2021.
\newblock \doi{10.1016/j.inffus.2021.05.008}.

\bibitem[Atan et~al.(2018)Atan, Zame, and van~der Schaar]{atan_lo18}
O.~Atan, W.R. Zame, and M.~van~der Schaar.
\newblock Learning optimal policies from observational data.
\newblock In \emph{Proc.\ ICML, 35th Int.\ Conf.\ on Machine Learning},
  Stockholm, Sweden, 2018.

\bibitem[Athey and Imbens(2016)]{athe_rp16}
S.~Athey and G.~Imbens.
\newblock Recursive partitioning for heterogeneous causal effects.
\newblock \emph{Proc.\ of the National Academy of Science}, 113\penalty0
  (27):\penalty0 7353--7360, 2016.

\bibitem[Bakir et~al.(2007)Bakir, Hofmann, Sch\"olkopf, Smola, Taskar, and
  Vishwanathan]{baki_ps}
G.~Bakir, T.~Hofmann, B.~Sch\"olkopf, A.~Smola, B.~Taskar, and S.~Vishwanathan,
  editors.
\newblock \emph{Predicting Structured Data}.
\newblock MIT Press, 2007.

\bibitem[Bengs and H\"ullermeier(2020)]{mpub420}
V.\ Bengs and E.~H\"ullermeier.
\newblock Preselection bandits.
\newblock In \emph{Proc.\ ICML, 37th Int.\ Conf.\ on Machine Learning}, Vienna,
  Austria, 2020.

\bibitem[Berger(1980)]{berg_sd80}
J.O. Berger.
\newblock \emph{Statistical Decision Theory: Foundations, Concepts, and
  Methods}.
\newblock Springer-Verlag, 1980.

\bibitem[Berger(1985)]{berg_sd}
J.O. Berger.
\newblock \emph{Statistical Decision Theory and Bayesian Analysis}.
\newblock Springer-Verlag, 1985.

\bibitem[Bertsimas and Kallus(2020)]{bert_tp20}
D.\ Bertsimas and N.\ Kallus.
\newblock From predictive to prescriptive analytics.
\newblock \emph{Management Science}, 66\penalty0 (3):\penalty0 1025--1044,
  2020.

\bibitem[Bertsimas et~al.(2019)Bertsimas, Dunn, and Mundru]{bert_op19}
D.~Bertsimas, J.~Dunn, and N.~Mundru.
\newblock Optimal prescriptive trees.
\newblock \emph{INFORMS Journal on Optimization}, 1\penalty0 (2):\penalty0
  164--183, 2019.
\newblock \doi{10.1287/ijoo.2018.0005}.

\bibitem[Beygelzimer and Langford(2009)]{beyg_to09}
A.~Beygelzimer and J.~Langford.
\newblock The offset tree for learning with partial labels.
\newblock In \emph{Proc.\ KDD, 15th ACM SIGKDD Int.\ Conf.\ on Knowledge
  Discovery and Data Mining}, pages 129--138, 2009.

\bibitem[Blaszczynski et~al.(2007)Blaszczynski, Greco, and
  Slowinski]{blas_mc07}
J.~Blaszczynski, S.~Greco, and R.~Slowinski.
\newblock Multi-criteria classification -- {A} new scheme for application of
  dominance-based decision rules.
\newblock \emph{European Journal of Operational Research}, 181\penalty0
  (3):\penalty0 1030--1044, 2007.

\bibitem[Bottou et~al.(2013)Bottou, Peters, Candela, Charles, Chickering,
  Portugaly, Ray, Simard, and Snelson]{bott_cr13}
L.~Bottou, J.~Peters, J.Q. Candela, D.X. Charles, M.~Chickering, E.~Portugaly,
  D.~Ray, P.Y. Simard, and E.~Snelson.
\newblock Counterfactual reasoning and learning systems: the example of
  computational advertising.
\newblock \emph{Journal of Machine Learning Research}, 14\penalty0
  (1):\penalty0 3207--3260, 2013.

\bibitem[Casteluccia and M\'etayer(2019)]{cast_ua19}
C.~Casteluccia and D.~Le M\'etayer.
\newblock Understanding algorithmic decision making: opportunities and
  challenges.
\newblock Technical Report PE 624.261, EPRS Study, European Parlament, 2019.

\bibitem[Chevaleyre et~al.(2008)Chevaleyre, Endriss, Lang, and
  Maudet]{chev_ph08}
Y.~Chevaleyre, U.~Endriss, J.~Lang, and N.~Maudet.
\newblock Preference handling in combinatorial domains: {F}rom {AI} to social
  choice.
\newblock \emph{AI Magazine}, 29\penalty0 (4), 2008.

\bibitem[Corrente et~al.(2013)Corrente, Greco, Kadzinski, and
  Slowinski]{corr_ro13}
S.~Corrente, S.~Greco, M.~Kadzinski, and R.~Slowinski.
\newblock Robust ordinal regression in preference learning and ranking.
\newblock \emph{Machine Learning}, 93\penalty0 (2-3):\penalty0 381--422, 2013.

\bibitem[Doppa et~al.(2014)Doppa, Fern, and Tadepalli]{dopp_sp14}
J.R. Doppa, A.~Fern, and P.~Tadepalli.
\newblock Structured prediction via output space search.
\newblock \emph{Journal of Machine Learning Research}, 15:\penalty0 1317--1350,
  2014.

\bibitem[Dragone et~al.(2018)Dragone, Teso, and Passerini]{drag_cp18}
P.\ Dragone, S.\ Teso, and A.\ Passerini.
\newblock Constructive preference elicitation over hybrid combinatorial spaces.
\newblock In \emph{Proc.\ AAAI Conf.\ on Artificial Intelligence}, 2018.

\bibitem[Elmachtoub et~al.(2020)Elmachtoub, Liang, and McNellis]{elma_dt20}
A.N. Elmachtoub, J.C.M. Liang, and R.~McNellis.
\newblock Decision trees for decision-making under the predict-then-optimize
  framework.
\newblock In \emph{Proc.\ ICML, 37th Int.\ Conf.\ on Machine Learning}, volume
  119 of \emph{PMLR}, 2020.

\bibitem[Ferguson(1967)]{ferg_ms}
T.~Ferguson.
\newblock \emph{Mathematical Statistics: A Decision Theoretic Approach}.
\newblock Academic Press, New York, 1967.

\bibitem[Fernandez and Provost(2019)]{fern_ov19}
C.~Fernandez and F.~Provost.
\newblock Observational vs experimental data when making automated decisions
  using machine learning.
\newblock Technical report, NYU Stern School of Business, 2019.
\newblock URL \url{SSRN: https://ssrn.com/abstract=3444678}.

\bibitem[Friedler et~al.(2021)Friedler, Scheidegger, and
  Venkatasubramanian]{frie_ti21}
S.A. Friedler, C.~Scheidegger, and S.~Venkatasubramanian.
\newblock The (im)possibility of fairness: Different value systems require
  different mechanisms for fair decision making.
\newblock \emph{Communications of the ACM}, 64\penalty0 (4):\penalty0 136--143,
  2021.
\newblock \doi{10.1145/3433949}.

\bibitem[F\"urnkranz and H\"ullermeier(2010)]{mpub218}
J.~F\"urnkranz and E.~H\"ullermeier, editors.
\newblock \emph{Preference Learning}.
\newblock Springer-Verlag, 2010.

\bibitem[Gambhir and Gupta(2017)]{gamb_ra17}
M.~Gambhir and V.~Gupta.
\newblock Recent automatic text summarization techniques: a survey.
\newblock \emph{Artificial Intelligence Review}, 47:\penalty0 1--66, 2017.
\newblock \doi{10.1007/s10462-016-9475-9}.

\bibitem[Goodman and Flaxman(2017)]{good_eu17}
R.\ Goodman and S.\ Flaxman.
\newblock European {U}nion regulations on algorithmic decision-making and a
  ``right to explanation''.
\newblock \emph{AI Magazine}, 38\penalty0 (3):\penalty0 1--9, 2017.

\bibitem[Grote and Berens(2020)]{grot_ot20}
T.~Grote and P.~Berens.
\newblock On the ethics of algorithmic decision-making in healthcare.
\newblock \emph{Journal of Medical Ethics}, 46\penalty0 (3):\penalty0 205--211,
  2020.
\newblock \doi{10.1136/medethics-2019-105586}.

\bibitem[Hansen(2020)]{hans_tv20}
K.B. Hansen.
\newblock The virtue of simplicity: {O}n machine learning models in algorithmic
  trading.
\newblock \emph{Big Data and Society}, January 2020.
\newblock \doi{10.1177/2053951720926558}.

\bibitem[Hartford et~al.(2017)Hartford, Lewis, Leyton-Brown, and
  Taddy]{hart_di17}
J.~Hartford, G.~Lewis, K.~Leyton-Brown, and M.~Taddy.
\newblock Deep {IV}: {A} flexible approach for counterfactual prediction.
\newblock In \emph{Proc.\ ICML, 34th Int.\ Conf.\ on Machine Learning},
  number~70 in PMLR, Sydney, Australia, 2017.

\bibitem[H\"ullermeier(2005)]{mpub065}
E.~H\"ullermeier.
\newblock Experience-based decision making: {A} satisficing decision tree
  approach.
\newblock \emph{IEEE Transactions on Systems, Man, and Cybernetics\,--\,Part A:
  Systems and Humans}, 35\penalty0 (5):\penalty0 641--653, 2005.

\bibitem[H\"ullermeier and Waegeman(2021)]{mpub440}
E.~H\"ullermeier and W.\ Waegeman.
\newblock Aleatoric and epistemic uncertainty in machine learning: {A}n
  introduction to concepts and methods.
\newblock \emph{Machine Learning}, 110\penalty0 (3):\penalty0 457--506, 2021.
\newblock \doi{10.1007/s10994-021-05946-3}.

\bibitem[Kallus(2017)]{kall_rp17}
N.~Kallus.
\newblock Recursive partitioning for personalization using observational data.
\newblock In \emph{Proc.\ ICML, 34th Int.\ Conf.\ on Machine Learning},
  volume~70 of \emph{PPMLR}, pages 1789--1798, 2017.

\bibitem[Kamiran et~al.(2012)Kamiran, Karim, and Zhang]{kami_dt12}
F.~Kamiran, A.~Karim, and X.~Zhang.
\newblock Decision theory for discrimination-aware classification.
\newblock In \emph{Proc.\ IEEE Int.\ Conf.\ on Data Mining}, pages 924--929,
  2012.
\newblock \doi{10.1109/ICDM.2012.45}.

\bibitem[Kendall and Gal(2017)]{kend_wu17}
A.\ Kendall and Y.\ Gal.
\newblock What uncertainties do we need in {B}ayesian deep learning for
  computer vision?
\newblock In \emph{Proc.\ NIPS, Advances in Neural Information Processing
  Systems}, pages 5574--5584, 2017.

\bibitem[Kleinberg et~al.(2018)Kleinberg, Lakkaraju, Leskovec, Ludwig, and
  Mullainathan]{klein_hd18}
J.~Kleinberg, H.~Lakkaraju, J.~Leskovec, J.~Ludwig, and S.~Mullainathan.
\newblock Human decisions and machine predictions.
\newblock \emph{The Quarterly Journal of Economics}, 133\penalty0 (1):\penalty0
  237--293, 2018.

\bibitem[Lattimore and Szepesvari(2020)]{latt_ba}
T.~Lattimore and C.~Szepesvari.
\newblock \emph{Bandit Algorithms}.
\newblock Cambridge University Press, 2020.

\bibitem[Lepri et~al.(2018)Lepri, Oliver, Letouz\'e, Pentland, and
  Vinck]{lepr_ft18}
B.~Lepri, N.~Oliver, E.~Letouz\'e, A.~Pentland, and P.~Vinck.
\newblock Fair, transparent, and accountable algorithmic decision-making
  processes.
\newblock \emph{Philosophy and Technology}, 31:\penalty0 611--627, 2018.
\newblock \doi{10.1007/s13347-017-0279-x}.

\bibitem[Maes et~al.(2009)Maes, Denoyer, and Gallinari]{maes_sp09}
F.\ Maes, L.\ Denoyer, and P.\ Gallinari.
\newblock Structured prediction with reinforcement learning.
\newblock \emph{Machine Learning}, 77\penalty0 (2--3):\penalty0 271--301, 2009.

\bibitem[Mannor and Tsitsiklis(2004)]{mann_ts04}
S.~Mannor and J.N. Tsitsiklis.
\newblock The sample complexity of exploration in the multi-armed bandit
  problem.
\newblock \emph{Journal of Machine Learning Research}, 5:\penalty0 623--648,
  2004.

\bibitem[Mehrabi et~al.(2021)Mehrabi, Morstatter, Saxena, Lerman, and
  Galstyan]{mehr_as21}
N.~Mehrabi, F.~Morstatter, N.~Saxena, K.~Lerman, and A.~Galstyan.
\newblock A survey on bias and fairness in machine learning.
\newblock \emph{ACM Computing Surveys}, 54\penalty0 (6):\penalty0 1--15, 2021.

\bibitem[Neumann and Morgenstern(1953)]{vonn_to}
J.~Von Neumann and O.~Morgenstern.
\newblock \emph{Theory of Games and Economic Behavior}.
\newblock John Wiley and Sons, 1953.

\bibitem[Pearl(2000)]{pear_cm}
J.~Pearl.
\newblock \emph{Causality: Models, Reasoning, and Inference}.
\newblock Cambridge University Press, 2000.

\bibitem[Pessach et~al.(2020)Pessach, Singer, Avrahamia, Ben-Gal, Shmueli, and
  Ben-Gala]{pess_er20}
D.~Pessach, G.~Singer, D.~Avrahamia, H.~Chalutz Ben-Gal, E.~Shmueli, and
  I.~Ben-Gala.
\newblock Employees recruitment: {A} prescriptive analytics approach via
  machine learning and mathematical programming.
\newblock \emph{Decision Support Systems}, 134, 2020.

\bibitem[Pitoura et~al.(2021)Pitoura, Stefanidis, and Koutrika]{pito_fi21}
E.~Pitoura, K.~Stefanidis, and G.~Koutrika.
\newblock Fairness in rankings and recommendations: {A}n overview.
\newblock \emph{CoRR}, abs/2104.05994v1, 2021.
\newblock URL \url{http://arxiv.org/abs/2104.05994v1}.

\bibitem[Powers et~al.(2017)Powers, Qian, Jung, Schuler, Nigam, Hastie, and
  Tibshirani]{powe_sm17}
S.~Powers, J.~Qian, K.~Jung, A.~Schuler, H.S. Nigam, T.~Hastie, and
  R.~Tibshirani.
\newblock Some methods for heterogenous treatment effect estimation in high
  dimensions.
\newblock Technical report, Stanford University, Palo Alto, CA, USA, 2017.

\bibitem[Prosperi et~al.(2020)Prosperi, Guo, Sperrin, Koopman, Min, He, Rich,
  Wang, Buchan, and Bian]{pros_ci20}
M.~Prosperi, Y.~Guo, M.~Sperrin, J.S. Koopman, J.S. Min, X.~He, S.~Rich,
  M.~Wang, I.E. Buchan, and J.~Bian.
\newblock Causal inference and counterfactual prediction in machine learning
  for actionable healthcare.
\newblock \emph{Nature Machine Intelligence}, 2:\penalty0 369--375, 2020.

\bibitem[Qian and Murphy(2011)]{qian_pg11}
M.~Qian and S.A. Murphy.
\newblock Performance guarantees for individualized treatment rules.
\newblock \emph{Annals of Statistics}, 39\penalty0 (2), 2011.

\bibitem[Reverdy et~al.(2017)Reverdy, Srivastava, and Leonard]{reve_si17}
P.~Reverdy, V.~Srivastava, and N.E. Leonard.
\newblock Satisficing in multi-armed bandit problems.
\newblock \emph{IEEE Transactions on Automatic Control}, 62\penalty0
  (8):\penalty0 3788--3803, 2017.

\bibitem[Rosenbaum and Rubin(1983)]{rose_tc83}
P.~Rosenbaum and D.~Rubin.
\newblock The central role of the propensity score in observational studies for
  causal effects.
\newblock \emph{Biometrika}, 70\penalty0 (1):\penalty0 41--55, 1983.

\bibitem[Russell(2016)]{russ_ra16}
S.J. Russell.
\newblock Rationality and intelligence: {A} brief update.
\newblock In V.~M\"uller, editor, \emph{Fundamental Issues of Artificial
  Intelligence}. Springer, Cham., 2016.

\bibitem[Savage(1954)]{sava_tf}
L.J. Savage.
\newblock \emph{The Foundations of Statistics}.
\newblock John Wiley and Sons, Inc., New York, 1954.

\bibitem[Senge et~al.(2014)Senge, B\"osner, Dembczynski, Haasenritter, Hirsch,
  Donner-Banzhoff, and H\"ullermeier]{mpub272}
R.\ Senge, S.\ B\"osner, K.\ Dembczynski, J.\ Haasenritter, O.\ Hirsch, N.\
  Donner-Banzhoff, and E.~H\"ullermeier.
\newblock Reliable classification: Learning classifiers that distinguish
  aleatoric and epistemic uncertainty.
\newblock \emph{Information Sciences}, 255:\penalty0 16--29, 2014.

\bibitem[Shrestha and Yang(2019)]{shre_fi19}
Y.R. Shrestha and Y.~Yang.
\newblock Fairness in algorithmic decision-making: {A}pplications in
  multi-winner voting, machine learning, and recommender systems.
\newblock \emph{Algorithms}, 12\penalty0 (199), 2019.
\newblock \doi{10.3390/a12090199}.

\bibitem[Slowik and Bottou(2021)]{slow_ab21}
A.~Slowik and L.~Bottou.
\newblock Algorithmic bias and data bias: Understanding the relation between
  distributionally robust optimization and data curation.
\newblock \emph{CoRR}, abs/2106.09467v1, 2021.
\newblock URL \url{http://arxiv.org/abs/2106.09467v1}.

\bibitem[Sutton and Barto(2018)]{sutt_rl}
R.S.\ Sutton and A.G.\ Barto.
\newblock \emph{Reinforcement Learning: {A}n Introduction}.
\newblock Bradford Books, 2 edition, 2018.

\bibitem[Swaminathan and Joachims(2015{\natexlab{a}})]{swam_cr15}
A.~Swaminathan and T.~Joachims.
\newblock Counterfactual risk minimization: {L}earning from logged bandit
  feedback.
\newblock In \emph{Proc.\ ICML, Int.\ Conf.\ on Machine Learning}, pages
  814--823, 2015{\natexlab{a}}.

\bibitem[Swaminathan and Joachims(2015{\natexlab{b}})]{swam_ts15}
A.~Swaminathan and T.~Joachims.
\newblock The self-normalized estimator for counterfactual learning.
\newblock In \emph{Proc. NIPS, Advances in Neural Information Processing
  Systems}, volume~28, 2015{\natexlab{b}}.

\bibitem[Tavakol(2020)]{tava_fc20}
M.~Tavakol.
\newblock Fair classification with counterfactual learning.
\newblock In \emph{Proc.\ SIGIR, 43rd Int.\ {ACM} Conf.\ on Research and
  Development in Information Retrieval}, pages 2073--2076, Virtual Event,
  China, 2020. {ACM}.
\newblock \doi{10.1145/3397271.3401291}.

\bibitem[Wager and Athey(2018)]{wage_ea18}
S.~Wager and S.~Athey.
\newblock Estimation and inference of heterogeneous treatment effects using
  random forests.
\newblock \emph{Journal of the American Statistical Association}, 113\penalty0
  (523):\penalty0 1228--1242, 2018.

\bibitem[Westreich et~al.(2010)Westreich, Lessler, and Funk]{west_ps10}
D.~Westreich, J.~Lessler, and M.J. Funk.
\newblock Propensity score estimation: {M}achine learning and classification
  methods as alternatives to logistic regression.
\newblock \emph{Journal of Clinical Epidemiology}, 63\penalty0 (8), 2010.

\bibitem[Wheeler(2020)]{whee_br20}
G.~Wheeler.
\newblock Bounded rationality.
\newblock In E.N. Zalta, editor, \emph{The {Stanford} Encyclopedia of
  Philosophy}. Metaphysics Research Lab, Stanford University, {F}all 2020
  edition, 2020.

\bibitem[Zagorecki et~al.(2013)Zagorecki, Johnson, and Ristvej]{zago_dm13}
A.T. Zagorecki, D.E.A. Johnson, and J.~Ristvej.
\newblock Data mining and machine learning in the context of disaster and
  crisis management.
\newblock \emph{Int.\ Journal of Emergency Management}, 9\penalty0 (4), 2013.

\bibitem[Zehlike et~al.(2021)Zehlike, Yang, and Stoyanovich]{zehl_fi21}
M.~Zehlike, K.~Yang, and J.~Stoyanovich.
\newblock Fairness in ranking: {A} survey.
\newblock \emph{CoRR}, abs/2103.14000v2, 2021.
\newblock URL \url{http://arxiv.org/abs/2103.14000v2}.

\bibitem[Zhao et~al.(2012)Zhao, Zeng, Rush, and Kosorok]{zhao_ei12}
Y.~Zhao, D.~Zeng, A.J. Rush, and M.R. Kosorok.
\newblock Estimating individualized treatment rules using outcome weighted
  learning.
\newblock \emph{Journal of the American Statistical Association}, 107\penalty0
  (499):\penalty0 1106--1118, 2012.
\newblock \doi{10.1080/01621459.2012.695674}.

\bibitem[Zhou et~al.(2017)Zhou, Mayer-Hamblett, Khan, and Kosorok]{zhou_rw17}
X.~Zhou, N.~Mayer-Hamblett, U.~Khan, and M.R. Kosorok.
\newblock Residual weighted learning for estimating individualized treatment
  rules.
\newblock \emph{Journal of the American Statistical Association}, 112\penalty0
  (517):\penalty0 169--187, 2017.

\bibitem[Zhou(2018)]{zhou_ab18}
Z.H. Zhou.
\newblock A brief introduction to weakly supervised learning.
\newblock \emph{National Science Review}, 5:\penalty0 44--53, 2018.

\bibitem[Zliobaite(2017)]{zlio_md17}
I.~Zliobaite.
\newblock Measuring discrimination in algorithmic decision making.
\newblock \emph{Data Mining and Knowledge Discovery}, 31:\penalty0 1060--1089,
  2017.
\newblock \doi{10.1007/s10618-017-0506-1}.

\end{thebibliography}

\end{document}